\documentclass[11pt,a4paper]{article}
\usepackage[hyperref]{naaclhlt2019}
\usepackage{times}
\usepackage{latexsym}
\usepackage{subcaption}
\usepackage{amsmath,amssymb}
\usepackage{tikz}
\usetikzlibrary{calc,trees,positioning,arrows,chains,shapes.geometric,%
    decorations.pathreplacing,decorations.pathmorphing,shapes,%
    matrix,shapes.symbols}

\tikzset{suppress join/.code={\def\tikz@after@path{}}}

\tikzset{
>=stealth',
  punktchain/.style={
    rectangle, 
    rounded corners, 
    draw=black, thin,
    text width=8em, 
    minimum height=3em, 
    text centered,
    on chain},
  circlechain/.style={
    circle,  
    draw=black, thin,
    text centered, 
    minimum size=3em,
    inner sep=1pt,
    on chain},
  gen_circle/.style={
    circle,  
    draw=black, thin,
    text centered, 
    minimum size=3em,
    inner sep=1pt,
    on chain},
  line/.style={draw, thick, <-},
  side/.style={
    rectangle,
    minimum width=1em,
    draw=black, thin,
    text width=2em, 
    minimum height=2em, 
    text centered},
  sidecircle/.style={
    circle,  
    fill=green!10,
    draw=black, thin,
    text centered, 
    minimum size=0.1em,
    inner sep=1pt,
    on chain},
  every join/.style={->, thin,shorten >=1pt},
  decoration={brace},
  tuborg/.style={decorate},
  tubnode/.style={midway, right=2pt},
  bignode/.style={font={\fontsize{23}{25}\selectfont}},
  regularnode/.style={font={\fontsize{18}{25}\selectfont}},
  smallnode/.style={font={\fontsize{10}{12}\selectfont}},
}
\usepackage{multirow}

\usepackage{url}
\usepackage{xspace}
\usepackage{bbm}
\aclfinalcopy 

\newcommand{\kld}{\mathit{KL}}
\newcommand{\vmf}{\text{vMF}}
\DeclareMathOperator*{\argmax}{argmax}

\newcommand{\vg}{VGVAE\xspace}
\newcommand{\wordavg}{\textsc{Wordavg}\xspace}
\newcommand{\lstmavg}{\textsc{BLSTMavg}\xspace}
\newcommand{\prl}{PRL\xspace}
\newcommand{\spl}{DPL\xspace}
\newcommand{\wpl}{WPL\xspace}

\title{A Multi-Task Approach for Disentangling Syntax and Semantics \\ in Sentence Representations}

\author{Mingda Chen\qquad Qingming Tang\qquad Sam Wiseman\qquad Kevin Gimpel\\
Toyota Technological Institute at Chicago, Chicago, IL, 60637, USA\\
  {\tt \{mchen,qmtang,swiseman,kgimpel\}@ttic.edu}\\}

\date{}

\begin{document}
\maketitle
\begin{abstract}
We propose a generative model for a sentence that uses two latent variables, with one intended to represent the syntax of the sentence and the other to represent its semantics. We show we can achieve better disentanglement between semantic and syntactic representations by training with multiple losses, including losses that exploit aligned paraphrastic sentences and word-order information. We also investigate the effect of moving from bag-of-words to recurrent neural network modules. We evaluate our models as well as several popular pretrained embeddings on standard semantic similarity tasks and novel syntactic similarity tasks. 
Empirically, we find that the model with the best performing syntactic and semantic representations also gives rise to the most disentangled representations.\footnote{Code and data are available at \href{https://github.com/mingdachen/disentangle-semantics-syntax}{\nolinkurl{github.com/mingdachen/disentangle-semantics-syntax}}}
\end{abstract}

\section{Introduction}
As generative latent variable models, especially of the continuous variety~\cite{Kingma2014,goodfellow2014generative}, have become increasingly important in natural language processing~\citep{bowman2016generating,Gulrajani2017wgan}, there has been increased interest in learning models where the latent representations are disentangled~\cite{hu17control}. Much of the recent NLP work on learning disentangled representations of text has focused on disentangling the representation of attributes such as sentiment from the representation of content, typically in an effort to better control text generation~\cite{shen2017style,zhao2017learning,fu2018style}.

In this work, we instead focus on learning sentence representations that disentangle the syntax and the semantics of a sentence. We are moreover interested in disentangling these representations not for the purpose of controlling generation, but for the purpose of calculating semantic or syntactic similarity between sentences (but not both). To this end, we propose a generative model of a sentence 
which makes use of both semantic and syntactic latent variables, and we evaluate the induced representations on both standard semantic similarity tasks and
on several novel syntactic similarity tasks. 

We use a deep generative model consisting of von Mises Fisher (vMF) and Gaussian priors on the semantic and syntactic latent variables (respectively) and a deep bag-of-words decoder that
conditions on these latent variables. Following much recent work, we learn this model by optimizing the ELBO with a VAE-like~\cite{Kingma2014,Rezende2014} approach.

Our learned semantic representations are evaluated on the SemEval semantic textual similarity (STS) tasks~\cite{agirre2012semeval,cer2017semeval}. Because there has been less work on evaluating syntactic representations of sentences, we propose several new syntactic evaluation tasks, which involve predicting the syntactic analysis of an unseen sentence to be the syntactic analysis of its nearest neighbor (as determined by the latent syntactic representation) in a large set of annotated sentences.

In order to improve the quality and disentanglement of the learned representations, we incorporate simple additional losses in our training, which are designed to force the latent representations to capture different information. In particular, our semantic multi-task losses make use of aligned paraphrase data, whereas our syntactic multi-task loss makes use of word-order information. Additionally, we explore different encoder and decoder architectures for learning better syntactic representations.
Experimentally, we find that by training in this way we are able to force the learned representations to capture different information (as measured by the performance gap between the latent representations on each task). Moreover, we find that we achieve the best performance on all tasks when the learned representations are most disentangled.

\section{Related Work}
There is a growing amount of work on learning interpretable or disentangled latent representations both in machine learning~\cite{tenenbaum2000separating,reed2014learning,Makhzani2015,mathieu2016disentangling,higgins2016beta,Chen16infogan,hsu17unsup} and in various NLP applications, including sentence sentiment and style transfer~\cite[\emph{inter alia}]{hu17control,shen2017style,fu2018style,zhao2018adversarially}, morphological reinflection~\cite{zhou2017multi}, semantic parsing~\cite{yin18struct}, text generation~\cite{wiseman2018templates}, and sequence labeling~\cite{chen2018vsl}. Another related thread of work is text-based variational autoencoders~\cite{miao2016neural,bowman2016generating,serban2017piecewise,xu2018spherical}.

In terms of syntax and semantics in particular, there is a rich history of work in analyzing their interplay in sentences~\citep{C88-1057,van2005exploring}. We do not intend to claim that the two can be entirely disentangled in distinct representations. Rather, our goal is to propose modica of knowledge via particular multi-task losses and measure the extent to which this knowledge leads learned representations to favor syntactic or semantic information from a sentence. 

There has been prior work with similar goals for representations of words~\citep{P15-1126} and bilexical dependencies~\citep{W16-1615}, finding that decomposing syntactic and semantic information can lead to improved performance on semantic tasks. We find similar trends in our results, but at the level of sentence representations. A similar idea has been explored for text generation~\cite{iyyer2018adversarial}, where adversarial examples are generated by controlling syntax.

Some of our losses use sentential paraphrases, relating them to
work in paraphrase modeling~\cite{wieting-16-full,para-nmt-acl-18}. 
\citet{deudon2018learning} recently proposed a variational framework for modeling paraphrastic sentences, but our focus here is on learning disentangled representations.

As part of our evaluation, we develop novel syntactic similarity tasks for sentence representations learned without any syntactic supervision. These evaluations relate to the broad range of work in unsupervised parsing~\citep{P04-1061} and part-of-speech tagging~\citep{D10-1056}. However, our evaluations differ from previous evaluations in that we employ $k$-nearest-neighbor syntactic analyzers using our syntactic representations to choose nearest neighbors. 

There is a great deal of work on applying multi-task learning to various NLP tasks~\cite[\emph{inter alia}]{plank2016multilingual,rei2017semi,augen2017multitask,bollmann18multitask} and, recently, as a way of improving the quality or disentanglement of learned representations~\cite{zhao2017learning,goyal2017z,du2018variational,john2018disentangled}.

\section{Proposed Approach}

\begin{figure}
    \centering
    \begin{tikzpicture}
  [node distance=1.5em,
  start chain=going right,
  scale=0.5,
  every node/.style={scale=0.5}]
	 \node[bignode, gen_circle] (x1) {$x$};
	 \node[right=2.0em of x1] (in) {};
	 \node[bignode, gen_circle, above=0.4em of in] (z) {$z$};
	 \node[bignode, gen_circle, below=0.4em of in] (y) {$y$};
	 \node[bignode, gen_circle, right=2.0em of in] (x2) {$x$};
	 
	 \draw[->,thin] (y.10) -> (x2.220);
	 \draw[->,thin] (z.350) -> (x2.140);

	 \draw[->,thin, dashed] (x1.50) -> (z.180);
	 \draw[->,thin, dashed] (x1.320) -> (y.160);
\end{tikzpicture}
    \caption{Graphical model of \vg. Dashed lines indicate inference model. Solid lines indicate generative model.}
    \label{fig:graph}
\end{figure}
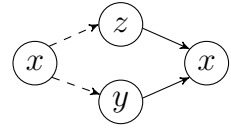

Our goal is to extract the disentangled semantic and syntactic information from sentence representations. To achieve this, we introduce the vMF-Gaussian Variational Autoencoder (\vg). As shown in Figure~\ref{fig:graph}, \vg assumes a sentence is generated by conditioning on two independent variables: semantic variable $y$ and syntactic variable $z$. In particular, our model gives rise to the following joint likelihood
\begin{align*}
    p_{\theta}(x, y, z) &= p_{\theta}(y) p_{\theta}(z) p_{\theta}(x \vert y, z) \\
    &= p_{\theta}(y) p_{\theta}(z) \prod_{t=1}^T p(x_t\, | \, y, z), 
\end{align*}
where $x_t$ is the $t$th word of $x$, $T$ is the sentence length, and $p(x_t \vert y, z)$ is given by a softmax over a vocabulary of size $V$. Further details on the parameterization are given below.

To perform inference, we assume a factored posterior $q_\phi(y,z\vert  x)=q_\phi(y\vert x)q_\phi(z\vert x)$, as has been used in prior work~\cite{zhou2017multi,chen2018vsl}. Learning of \vg maximizes a lower bound on marginal log-likelihood:

\begin{equation}
\begin{aligned}
    &\log p_\theta(x)\geq\mathop\mathbb{E}_{\substack{y\sim q_\phi(y\vert  x)\\z\sim q_\phi(z\vert  x)}}[\log p_\theta(x\vert z,y)\\
    &-\log\frac{q_\phi(z\vert  x)}{p_\theta(z)}
    -\log\frac{q_\phi(y\vert  x)}{p_\theta(y)}]\\
    &=\mathop\mathbb{E}_{\substack{y\sim q_\phi(y\vert  x)\\z\sim q_\phi(z\vert  x)}}[\log p_\theta(x\vert  z,y)]-\kld(q_\phi(z\vert  x)\Vert p_\theta(z))\\
    &-\kld(q_\phi(y\vert  x)\Vert p_\theta(y))\stackrel{\text{def}}{=\joinrel=}\text{ELBO}
\end{aligned}
\label{eq:elbo}
\end{equation}

\subsection{Parameterizations}
\vg uses two distribution families in defining the posterior over latent variables, namely, the von Mises-Fisher (vMF) distribution and the Gaussian distribution.

\paragraph{vMF Distribution.} vMF can be regarded as a Gaussian distribution on a hypersphere with two parameters: $\mu$ and $\kappa$. $\mu\in\mathbb{R}^m$ is a normalized vector (i.e. $\Vert\mu\Vert_2=1$ ) defining the mean direction. $\kappa\in\mathbb{R}_{\geq 0}$ is often referred to as a concentration parameter analogous to the variance in a Gaussian distribution. vMF has been used for modeling similarity between two sentences~\cite{kelvin2018gen}, which is particularly suited to our purpose here, since we will evaluate our semantic representations in the context of modeling paraphrases (See Sections~\ref{para-recon-loss} and \ref{disc-para-loss} for more details).

Therefore, we assume $q_\phi(y\vert  x)$ follows $\vmf(\mu_\alpha(x),\kappa_\alpha(x))$ and the prior $p_\theta(y)$ follows the uniform distribution $\vmf(\cdot,0)$.

With this choice of prior and posterior distribution, the $\kld(q_\phi(y\vert  x)\Vert p_\theta(y))$ appearing in the ELBO can be computed in closed-form:

\begin{equation}
\begin{aligned}
    &\kappa_\alpha\frac{\mathcal{I}_{m/2}(\kappa_\alpha)}{\mathcal{I}_{m/2-1}(\kappa_\alpha)} + (m/2 - 1)\log\kappa_\alpha - \\
    &(m/2)\log(2\pi) - \log\mathcal{I}_{m/2-1}(\kappa_\alpha)+\\
    &\frac{m}{2}\log\pi+\log 2-\log\Gamma(\frac{m}{2}),
\end{aligned}
\end{equation}

\noindent where $\mathcal{I}_v$ is the modified Bessel function of the first kind at order $v$ and $\Gamma(\cdot)$ is the Gamma function. We follow~\citet{davidson2018hyperspherical} and use an acceptance-rejection scheme to sample from vMF.

\paragraph{Gaussian Distribution.\footnote{In preliminary experiments, we observed that using two distribution families can lead to better performance. This is presumably because the Gaussian distribution complements the norm information lost in the vMF distribution.}}

We assume $q_\phi(z\vert  x)$ follows a Gaussian distribution $\mathcal{N}(\mu_\beta(x),\text{diag}(\sigma_\beta(x)))$ and that the prior $p_\theta(z)$ is $\mathcal{N}(0,I_{d})$, where $I_{d}$ is an $d\times d$ identity matrix.

Since we only consider a diagonal covariance matrix, the KL divergence term $\kld(q_\phi(z\vert  x)\Vert p_\theta(z))$ can also be computed efficiently:

\begin{equation}
    \frac{1}{2}(-\sum_i\log\sigma_{\beta i} + \sum_i\sigma_{\beta i} + \sum_i{\mu_{\beta i} ^2} - d)
\end{equation}

\paragraph{Inference and Generative Models.}

The inference models $q_\phi(y\vert x)$ and $q_\phi(z\vert  x)$ are two independent word averaging encoders with additional linear feedforward neural networks for producing $\mu(x)$ and $\sigma(x)$ (or $\kappa(x)$). 

The generative model $p_\theta(x\vert  y,z)$ is a feedforward neural network $g_\theta$ with the output being a bag of words. In particular, the expected output log-probability (the first term in Eq.~\ref{eq:elbo}) is computed as follows:

\begin{equation}
\begin{aligned}
    &\mathop\mathbb{E}_{\substack{y\sim q_\phi(y\vert  x)\\z\sim q_\phi(z\vert  x)}}[\log p_\theta(x\vert  y,z)]=\\
    &\mathop\mathbb{E}_{\substack{y\sim q_\phi(y\vert  x)\\z\sim q_\phi(z\vert  x)}}\left[\sum_{t=1}^T \log \frac{\exp{g_\theta([y;z])}_{x_t}}{\sum_{j=1}^{V}\exp{g_\theta([y;z])}_j}\right]\nonumber
\end{aligned}
\end{equation}
\noindent Where $V$ is the vocabulary size, $[;]$ indicates concatenation, $T$ is the sentence length and $x_t$ is the index of the $t$'th word's word type.

\paragraph{Recurrent Neural Networks.}

\begin{figure}
    \centering
    \includegraphics[scale=0.2]{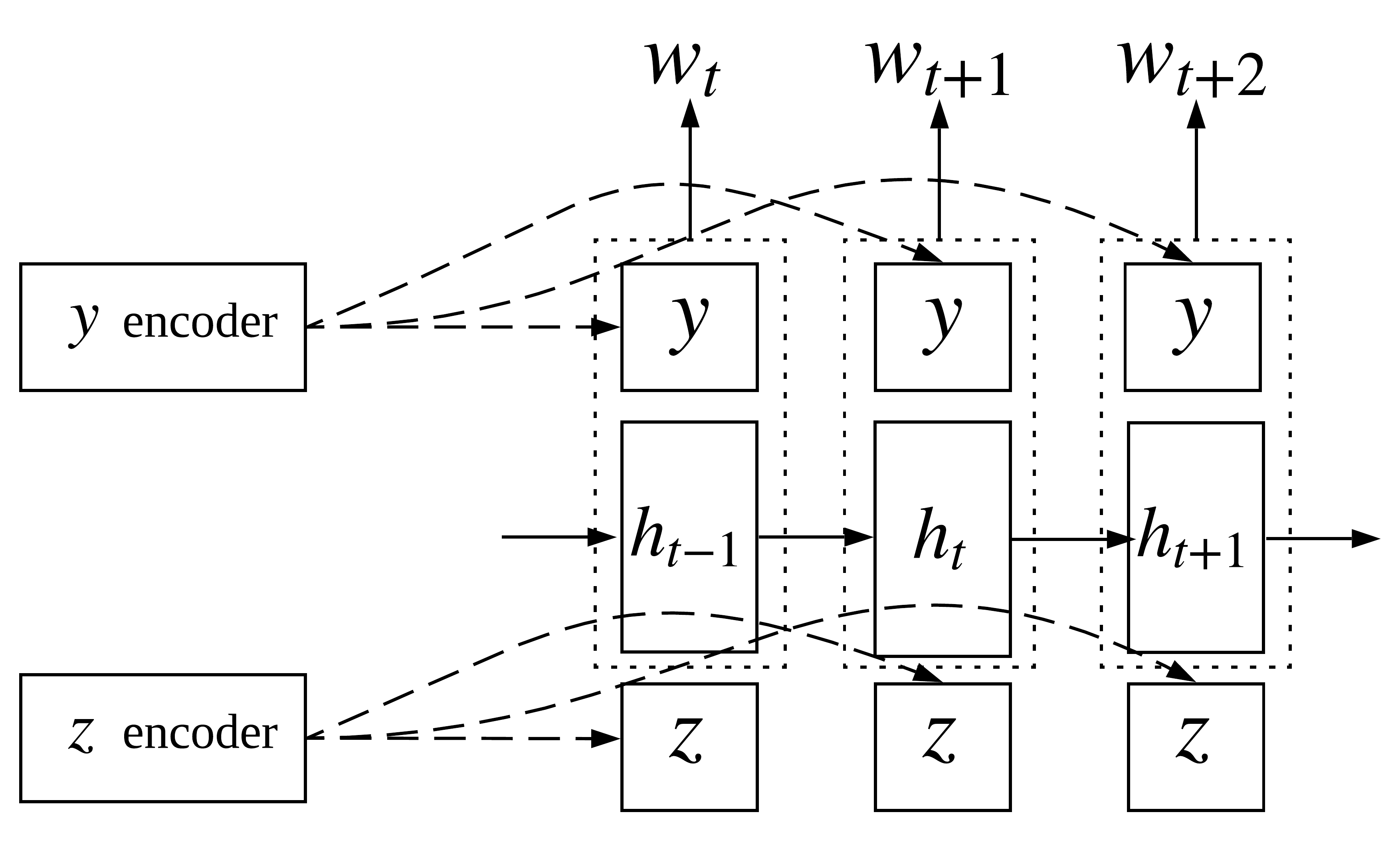}
    \caption{Diagram showing LSTM decoder that uses the semantic variable $y$ and the syntactic variable $z$.}
    \label{fig:vgvae-decoder}
\end{figure}

To facilitate better learning of syntax, we also consider replacing both the generative and inference models with RNN-based sequence models, rather than bag-of-words models. In this setting, the generative model $p_\theta(x\vert y,z)$ is a unidirectional long-short term memory network~(LSTM; \citealp{hochreiter1997long}) and a linear feedforward neural network for predicting the word tokens (shown in Figure~\ref{fig:vgvae-decoder}). The expected output log-probability is computed as follows:

\begin{equation}
\begin{aligned}
    &\mathop\mathbb{E}_{\substack{y\sim q_\phi(y\vert  x)\\z\sim q_\phi(z\vert  x)}}[\log p_\theta(x\vert  y,z)]=\\
    &\mathop\mathbb{E}_{\substack{y\sim q_\phi(y\vert  x)\\z\sim q_\phi(z\vert  x)}}\left[\sum_{t=1}^T \log p_\theta(x_t\vert y,z,x_{1:t-1})\right]\nonumber
\end{aligned}
\end{equation}

\noindent Where $V$ is the vocabulary size, $T$ is the sentence length and $x_t$ is the index of the $t$'th word's word type.

The inference model $q_\phi(y\vert x)$ is still a word averaging encoder, but $q_\phi(z \vert x)$ is parameterized by a bidirectional LSTM, where we concatenate the forward and backward hidden states and then take the average. The output of the LSTM is then used as input to a feedforward network with one hidden layer for producing $\mu(x)$ and $\sigma(x)$ (or $\kappa(x)$).

In the following sections, we will introduce several losses that will be added into the training of our base model, which empirically shows the ability of further disentangling the functionality between the semantic variable $y$ and the syntactic variable $z$.

\section{Multi-Task Training}
\begin{figure}
    \centering
    \begin{tikzpicture}
  [node distance=1.5em,
  start chain=going right,
  scale=0.5,
  every node/.style={scale=0.5}]
 \node[regularnode, punktchain] (e1) {$z$ encoder};
 \node[bignode, gen_circle, right=2.0em of e1] (z) {$z_1$};
 \node[bignode, gen_circle, left=1.0em of e1] (x1) {$x_1$};
 \node[regularnode, punktchain, below=2.0em of e1] (e2) {$y$ encoder};
 \node[bignode, gen_circle, right=2.0em of e2] (y) {$y_1$};
 \node[bignode, gen_circle, left=1.0em of e2] (x2) {$x_1$};
 
 \node[regularnode, punktchain, below=2.0em of e2] (e3) {$y$ encoder};
 \node[bignode, gen_circle, right=2.0em of e3] (z2) {$y_2$};
 \node[bignode, gen_circle, left=1.0em of e3] (x4) {$x_2$};
 \node[regularnode, punktchain, below=2.0em of e3] (e4) {$z$ encoder};
 \node[bignode, gen_circle, right=2.0em of e4] (y2) {$z_2$};
 \node[bignode, gen_circle, left=1.0em of e4] (x5) {$x_2$};

 \node[below=1.0em of e1] (middle) {};
 \node[bignode, gen_circle, right=6.5em of middle] (x3) {$x_1$};
 
 \draw[->,thin, dashed] (e1.0) -> (z.180);
 \draw[->,thin, dashed] (x1.0) -> (e1.180);
 
 \draw[->,thin, dashed] (e2.0) -> (y.180);
 \draw[->,thin, dashed] (x2.0) -> (e2.180);
 
 \%draw[->,thin] (z.350) -> (x3.135);
 
 \node[below=1.0em of e3] (middle) {};
 \node[bignode, gen_circle, right=6.5em of middle] (x6) {$x_2$};
 
 \draw[->,thin, dashed] (e3.0) -> (z2.180);
 \draw[->,thin, dashed] (x4.0) -> (e3.180);
 
 \draw[->,thin, dashed] (e4.0) -> (y2.180);
 \draw[->,thin, dashed] (x5.0) -> (e4.180);
 
 


 \node[below=0.8em of e2] (dpl_pos) {};
 \node[bignode, below=0.4em of e2] (dpl_des) {DPL};
 \node[rectangle, rounded corners,
 	   draw=black, thick, minimum width=5em,
 	   minimum height=13em, right=3.4em of dpl_pos, dashed] (dpl) {};

 \draw[->,thin, dotted] (dpl_des.east) -> (dpl.178);

 \draw[->,thin, dash dot] (y2.10) -> (x6.220);
 \draw[->,thin, dash dot] (z2.30) -> (x3.245);
 
 \draw[->,thin, dash dot] (y.330) -> (x6.120);
 \draw[->,thin, dash dot] (z.350) -> (x3.135);
\end{tikzpicture}
    \caption{Diagram showing the training process when using the discriminative paraphrase loss (\spl; dotted lines) and paraphrase reconstruction loss (\prl; dash-dotted lines). The pair $(x_1, x_2)$ is a sentential paraphrase pair, the $y$'s are the semantic variables corresponding to each $x$, and the $z$'s are syntactic variables. 
    }
    \label{fig:losses}
\end{figure}
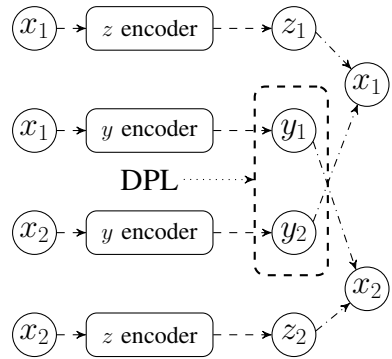

We attempt to improve the quality and disentanglement of our semantic and syntactic representations by introducing additional losses, which encourage $y$ to capture semantic information and $z$ to capture syntactic information. We elaborate on these losses below.

\subsection{Paraphrase Reconstruction Loss}
\label{para-recon-loss}
Our first loss is a paraphrase reconstruction loss (\prl). 
The key assumption underlying the PRL is that for a paraphrase pair $x_1$, $x_2$, the semantic information is equivalent between the two sentences and only the syntactic information varies. To impose such constraints, \prl is defined as

\begin{equation}
\begin{aligned}
    \mathop\mathbb{E}_{\substack{y_2\sim q_\phi(y\vert x_2)\\z_1\sim q_\phi(z\vert x_1)}}[&-\log p_\theta(x_1\vert y_2,z_1)] +\\ \mathop\mathbb{E}_{\substack{y_1\sim q_\phi(y\vert x_1)\\z_2\sim q_\phi(z\vert x_2)}}[&-\log p_\theta(x_2\vert y_1,z_2)]
\end{aligned}
\label{eqn:prl}
\end{equation}

That is, we swap the semantic variables, keep the syntactic variables, and attempt to reconstruct the sentences (shown in Figure~\ref{fig:losses}). 
While instead of using a multi-task objective we could directly model paraphrases $x_1$ and $x_2$ as being generated by the same $y$ (which naturally suggests a product-of-experts style posterior, as in \citet{wu2018multimodal}), we found that for the purposes of our downstream tasks training with the multi-task loss gave superior results. 

\subsection{Discriminative Paraphrase Loss}
\label{disc-para-loss}

Our second loss is a discriminative paraphrase loss (\spl). The \spl explicitly encourages the similarity of paraphrases $x_1$, $x_2$ to be scored higher than the dissimilar sentences $n_1$, $n_2$ (i.e., negative samples; see Sec.~\ref{neg-sample} for more details) by a given margin $\delta$. As shown in Figure~\ref{fig:losses}, the similarity function in this loss only uses the semantic variables in the sentences. The loss is defined as

\begin{equation}
\begin{aligned}
    &\max(0,\delta-d(x_1,x_2)+d(x_1,n_1))+\\
    &\max(0,\delta-d(x_1,x_2)+d(x_2,n_2))
\end{aligned}
\end{equation}
\noindent The similarity function we choose is the cosine similarity between the mean directions of the semantic variables from the two sentences: 
\begin{equation}
    d(x_1, x_2)=\mathrm{cosine}(\mu_\alpha(x_1), \mu_\alpha(x_2))
\end{equation}

\subsection{Word Position Loss}
It has been observed in previous work that word order typically contributes little to the modelling of semantic similarity~\citep{wieting-16-full}. We interpret this as evidence that word position information is more relevant to syntax than semantics, at least as evaluated by STS tasks. To guide the syntactic variable to represent word order, we introduce a word position loss (\wpl). Although our word averaging encoders 
only have access to the bag of words of the input, using this loss can be viewed as a denoising autoencoder where we have maximal input noise (i.e., an orderless representation of the input) and the encoders need to learn to reconstruct the ordering.

For both word averaging encoders and LSTM encoders, \wpl is parameterized by a three-layer feedforward neural network $f(\cdot)$ with input from the concatenation of the samples of the syntactic variable $z$
and the embedding vector $e_i$ at input position $i$; we then attempt to predict a one-hot vector representing the position $i$. More specifically, we define

\begin{equation}
    \text{WPL}\stackrel{\text{def}}{=\joinrel=}\mathop\mathbb{E}_{z\sim q_\phi(z\vert x)}\left[-\sum_{i}\log\textrm{softmax}(f([e_i;z]))_i\right]
    \nonumber
\end{equation}
\noindent where $\textrm{softmax}(\cdot)_i$ indicates the probability at position $i$.

\section{Training}
\paragraph{KL Weight.}
Following previous work on VAEs~\cite{higgins2016beta,alemi2016deep}, we attach a weight to the KL divergence and tune it based on development set performance.

\paragraph{Negative Samples.}
\label{neg-sample}
When applying \spl, we select negative samples based on maximizing cosine similarity to sentences from a subset of the data. In particular, we accumulate $k$ mini-batches during training, yielding a ``mega-batch'' $\mathcal{S}$ \citep{para-nmt-acl-18}. Then the negative samples are selected based on the following criterion:
\begin{equation}
    n_1=\argmax_{n\in\mathcal{S}\wedge n\neq x_2} \mathrm{cosine}(\mu_\alpha(x_1),\mu_\alpha(n))\nonumber
\end{equation}
\noindent where $x_1$, $x_2$ forms the paraphrase pair and the mega-batch size is fixed to $k=20$ for all of our experiments. Since all of our models are trained from scratch, we observed some instabilities with \spl during the initial stages of training. We suspect that this is because the negative samples at these initial stages are of low quality. To overcome this issue, \spl is included starting at the second epoch of training so that the models can have a warm start.

\section{Experiments}
\subsection{Setup}

We subsampled half a million paraphrase pairs from ParaNMT-50M~\cite{para-nmt-acl-18} as our training set. We use SemEval semantic textual similarity (STS) task 2017~\cite{cer2017semeval} as a development set. For semantic similarity evaluation, we use the STS tasks from 2012 to 2016~\cite{agirre2012semeval,diab2013eneko,agirre2014semeval,agirre2015semeval,agirre2016semeval} and the STS benchmark test set~\cite{cer2017semeval}. For evaluating syntactic similarity, we propose several evaluations. One uses the gold parse trees from the Penn Treebank~\cite{Marcus:1993:BLA:972470.972475}, and the others are based on automatically tagging and parsing five million paraphrases from ParaNMT-50M; we describe these tasks in detail below.

For hyperparameters, the dimensions of the latent variables are 50. The dimensions of word embeddings are 50. We use cosine similarity as similarity metric for all of our experiments. We tune the weights for \prl and reconstruction loss from 0.1 to 1 in increments of 0.1 based on the development set performance. 
We use one sample from each latent variable during training. When evaluating \vg based models on STS tasks, we use the mean direction of the semantic variable $y$, while for syntactic similarity tasks, we use the mean vector of the syntactic variable $z$.

\subsection{Baselines}

Our baselines are a simple word averaging (\wordavg) model and bidirectional LSTM averaging (\lstmavg) model, both of which have been shown to be very competitive for modeling semantic similarity when trained on paraphrases~\cite{para-nmt-acl-18}. Specifically, \wordavg takes the average over the word embeddings in the input sequence to obtain the sentence representation. \lstmavg uses the averaged hidden states of a bidirectional LSTM as the sentence representation, where forward and backward hidden states are concatenated. These models use 50 dimensional word embeddings and 50 dimensional LSTM hidden vectors per direction. These baselines are trained with \spl only. Additionally, we scramble the input sentence for \lstmavg since it has been reported beneficial for its performance in semantic similarity tasks~\citep{wieting-gimpel:2017:Long}. 

We also benchmark several pretrained embeddings on both semantic similarity and syntactic similarity datasets, including GloVe~\cite{pennington2014glove},\footnote{We use 300 dimensional Common Crawl  embeddings available at \href{https://nlp.stanford.edu/projects/glove}{\nolinkurl{nlp.stanford.edu/projects/glove}}} SkipThought~\cite{skipthoughts},\footnote{\href{https://github.com/ryankiros/skip-thoughts}{\nolinkurl{github.com/ryankiros/skip-thoughts}}} 
InferSent~\cite{infersent},\footnote{We use model V1 available at \href{https://github.com/facebookresearch/InferSent}{\nolinkurl{github.com/facebookresearch/InferSent}}} 
ELMo~\cite{N18-1202},\footnote{We use the original model available at \href{https://allennlp.org/elmo}{\nolinkurl{allennlp.org/elmo}}} and 
BERT~\cite{devlin2018bert}.\footnote{We use bert-large-uncased available at \href{https://github.com/huggingface/pytorch-pretrained-BERT}{\nolinkurl{github.com/huggingface/pytorch-pretrained-BERT}}}  
For GloVe, we average word embeddings to form sentence embeddings. For ELMo, we average the hidden states from three layers and then average the hidden states across time steps. For BERT, we use the averaged hidden states from the last attention block.

\section{Results and Analysis}
\subsection{Semantic Similarity}

\begin{table}[t]
\setlength{\tabcolsep}{4pt}
\centering
\small
\begin{tabular}{l|c|c|c|c}
        & \multicolumn{2}{c|}{semantic var.}& \multicolumn{2}{c}{syntactic var.}\\
        & bm & avg &  bm & avg \\
\hline
GloVe & 39.0 & 48.7 & - & - \\
SkipThought  & 42.1 & 42.0 & - & - \\
InferSent & 67.8 & 61.0 & - & - \\
ELMo & 57.7 & 60.3 & - & - \\
BERT & 4.5 & 15.0 & - & - \\
\hline
\wordavg & 71.9 & 64.8 &  -   & - \\
\lstmavg & 71.4 & 64.4 &  -   & - \\
\hline
\vg             & 45.5 & 42.7 & 40.8 & 43.2 \\
\vg+ \wpl          & 51.5 & 49.3 & 28.1 & 31.0 \\
\vg+ \spl         & 68.4 & 58.2 & 37.8 & 40.5 \\
\vg+ \prl          & 67.9 & 57.8 & 29.6 & 32.7 \\
\vg+ \prl+ \wpl       & 69.8 & 61.3 & 23.2 & 27.9 \\
\vg+ \prl+ \spl       & 71.2 & 64.2 & 31.7 & 33.9 \\
\vg+ \spl+ \wpl       & 71.0 & 63.5 & 24.1 & 29.0 \\
ALL    & 72.3 & 65.1 & 20.1 & 24.2 \\
ALL + LSTM enc. & 72.5 & 65.1 & 16.3 & 24.5 \\
ALL + LSTM enc. \& dec. & \textbf{72.9} &  \textbf{65.5} & \textbf{11.3} & \textbf{19.3}

\end{tabular}
\caption{Pearson correlation (\%) for STS test sets. bm: STS benchmark test set. avg: the average of Pearson correlation for each domain in the STS test sets from 2012 to 2016. Results are in bold if they are highest in the ``semantic variable'' columns or lowest in the ``syntactic variable'' columns. ``ALL'' indicates all of the multi-task losses are used.} 
\label{sts-res}
\end{table}

As shown in Table~\ref{sts-res}, the semantic and syntactic variables of our base \vg model show similar performance on the STS test sets. As we begin adding multi-task losses, however, the performance of these two variables gradually diverges, indicating that different information is being captured in the two variables. More interestingly, note that when \textit{any} of the three losses is added to the base \vg model (even the WPL loss which makes no use of paraphrases), the performance of the semantic variable increases and the performance of the syntactic variable decreases; this suggests that each loss is useful in encouraging the latent variables to learn complementary information. 

Indeed, the trend of additional losses both increasing semantic performance and decreasing syntactic performance holds even as we use more than two losses, except for the single case of \vg + PRL + DPL, where the syntactic performance increases slightly. Finally, we see that when the bag-of-words \vg model is used with all of the multi-task losses (``ALL''), we observe a large gap between the performance of the semantic and syntactic latent variables, as well as strong performance on the STS tasks that
outperforms all baselines. 

Using LSTM modules further strengthens the disentanglement between the two variables and leads to even better semantic performance. While using an LSTM encoder and a bag-of-words decoder
is difficult to justify from a generative modeling perspective, we include results with this configuration to separate out the contributions of the LSTM encoder and decoder.

\subsection{Syntactic Similarity}
So far, we have only confirmed empirically that the syntactic variable has learned to \textit{not} capture semantic information. To investigate what the syntactic variable has captured, we propose several syntactic similarity tasks.

\begin{table*}[t]
\setlength{\tabcolsep}{4pt}
\centering
\small
\begin{tabular}{l|c|c|c|c|c|c}
        & \multicolumn{2}{c|}{Constituent Parsing (TED, $\downarrow$)} & \multicolumn{2}{c|}{Constituent Parsing ($F_1$, $\uparrow$)} & \multicolumn{2}{c}{POS Tagging (\%Acc., $\uparrow$)}
        \\ 
\hline
GloVe & \multicolumn{2}{c|}{120.8} & \multicolumn{2}{c|}{27.3} & \multicolumn{2}{c}{23.9}  \\
SkipThought &  \multicolumn{2}{c|}{99.5} & \multicolumn{2}{c|}{30.9} & \multicolumn{2}{c}{29.6}  \\
InferSent &  \multicolumn{2}{c|}{138.9} & \multicolumn{2}{c|}{28.0} & \multicolumn{2}{c}{25.1}  \\
ELMo  &  \multicolumn{2}{c|}{103.8} & \multicolumn{2}{c|}{30.4} & \multicolumn{2}{c}{27.8}  \\
BERT  & \multicolumn{2}{c|}{101.7} & \multicolumn{2}{c|}{28.6} & \multicolumn{2}{c}{25.4}  \\
\hline
Random baseline & \multicolumn{2}{c|}{121.4} & \multicolumn{2}{c|}{19.2} & \multicolumn{2}{c}{12.9} \\
Upper bound performance & \multicolumn{2}{c|}{51.6} & \multicolumn{2}{c|}{71.1} & \multicolumn{2}{c}{62.3} \\
\hline
\wordavg   & \multicolumn{2}{c|}{107.0} & \multicolumn{2}{c|}{25.5} & \multicolumn{2}{c}{21.4} \\
\lstmavg   & \multicolumn{2}{c|}{106.8} & \multicolumn{2}{c|}{25.7} & \multicolumn{2}{c}{21.6} \\
\hline
        & semantic var.& syntactic var.& semantic var.& syntactic var.& semantic var.& syntactic var. \\
\hline
\vg                   & 109.3 & 111.4 & 25.2 & 25.0 & 21.1 & 21.0 \\
\vg+ \wpl             & 112.3 & 105.9 & \textbf{24.1} & 28.2 & \textbf{20.3} & 24.2 \\
\vg+ \spl             & 108.1 & 110.6 & 25.1 & 26.1 & 21.3 & 21.8 \\
\vg+ \prl             & 111.9 & 110.9 & 24.7 & 26.9 & 21.0 & 22.2 \\
\vg+ \spl+ \wpl       & 111.2 & 105.0 & 25.1 & 28.8 & 21.5 & 24.6 \\
\vg+ \prl+ \spl       & 108.0 & 110.4 & 25.0 & 26.2 & 21.1 & 22.1 \\
\vg+ \prl+ \wpl       & 109.4 & 105.1 & 24.4 & 28.1 & 20.6 & 23.6 \\
ALL                   & 110.0 & 104.7 & 25.4 & 29.3 & 21.4 & 25.5  \\
ALL + LSTM enc.       & 112.0 & 101.0 & 25.7 & 37.3 & 22.1  & 34.0 \\
ALL + LSTM enc. \& dec. & \textbf{114.6} & \textbf{100.5} & 25.3 & \textbf{38.8} & 21.4 & \textbf{35.7}
\end{tabular}
\caption{Syntactic similarity evaluations, showing tree edit distance (TED) and labeled $F_1$ score for constituent parsing, and accuracy (\%) for part-of-speech tagging. Numbers are bolded if they are worst in the ``semantic variable'' column or best in the ``syntactic variable'' column. ``ALL'' indicates all the multi-task losses are used.}
\label{tot-syntax-res}
\end{table*}

In particular, we consider using the syntactic latent variable in calculating nearest neighbors for a 1-nearest-neighbor syntactic parser or part-of-speech tagger. We use our latent variables to define the similarity function in these settings and evaluate the quality of the output parses and tag sequences using several metrics. 

Our first evaluation involves constituency parsing, and we use the standard training and test splits from the Penn Treebank.
We predict a parse tree for each sentence in the test set by finding its nearest neighbor in the training set based on the cosine similarity of the mean vectors for the syntactic variables. The parse tree of the nearest neighbor will then be treated as our prediction for the test sentence. Since the train and test sentences may differ in length, standard parse evaluation metrics are not applicable, so we use tree edit 
distance~\cite{zhang1989simple}\footnote{\href{https://github.com/timtadh/zhang-shasha}{\nolinkurl{github.com/timtadh/zhang-shasha}}} 
to compute the distance between two parse tree without considering word tokens.

To better understand the difficulty of this task, we introduce two baselines. The first randomly selects a training sentence. We calculate its performance by running it ten times and then reporting the average.
We also report the upper bound performance given the training set. Since computing tree edit distance is time consuming, we subsample 100 test instances and compute the minimum tree edit distance for each sampled instance.
Thus, this number can be seen as the approximated upper bound performance for this task given the training set.

To use a more standard metric for these syntactic similarity tasks, we must be able to retrieve training examples with the same number of words as the sentence we are trying to parse. We accordingly parse and tag the five million paraphrase subset of the ParaNMT training data using  Stanford CoreNLP~\citep{manning2014stanford}. To form a test set, we group sentences in terms of sentence length and subsample 300 sentences for each sentence length. After removing the paraphrases of the sentences in the test set, we use the rest of the training set as candidate sentences for nearest neighbor search, and we restrict nearest neighbors to have the same sentence length as the sentence we are attempting to parse or tag, which allows us to use standard metrics like labeled $F_1$ score and tagging accuracy for evaluation.

\subsubsection{Results}

As shown in Table~\ref{tot-syntax-res}, the syntactic variables and semantic variables demonstrate similar trends across these three syntactic tasks. Interestingly, both \spl and \prl help to improve the performance of the syntactic variables, even though these two losses are only imposed on the semantic variables. We saw an analogous pattern in Table~\ref{sts-res}, which again suggests that by pushing the semantic variables to learn information shared by paraphrastic sentences, we also encourage the syntactic variables to capture complementary syntactic information. 
We also find that adding \wpl brings the largest improvement to the syntactic variable, and keeps the syntactic information carried by the semantic variables at a relatively low level. Finally, when adding all three losses, the syntactic variable shows the strongest performance across the three tasks.

In addition, we observe that the use of the LSTM encoder improves syntactic performance by a large margin and the LSTM decoder improves further, which suggests that the use of the LSTM decoder contributes to the amount of syntactic information represented in the syntactic variable. 

\hyphenation{Skip-Thought}
Among pretrained representations, SkipThought shows the strongest performance overall and ELMo has the second best performance in the last two columns. While InferSent performs worst in the first column, it gives reasonable performance for the other two. BERT performs relatively well in the first column but worse in the other two. 

\begin{figure}
    \centering
    \includegraphics[scale=0.3]{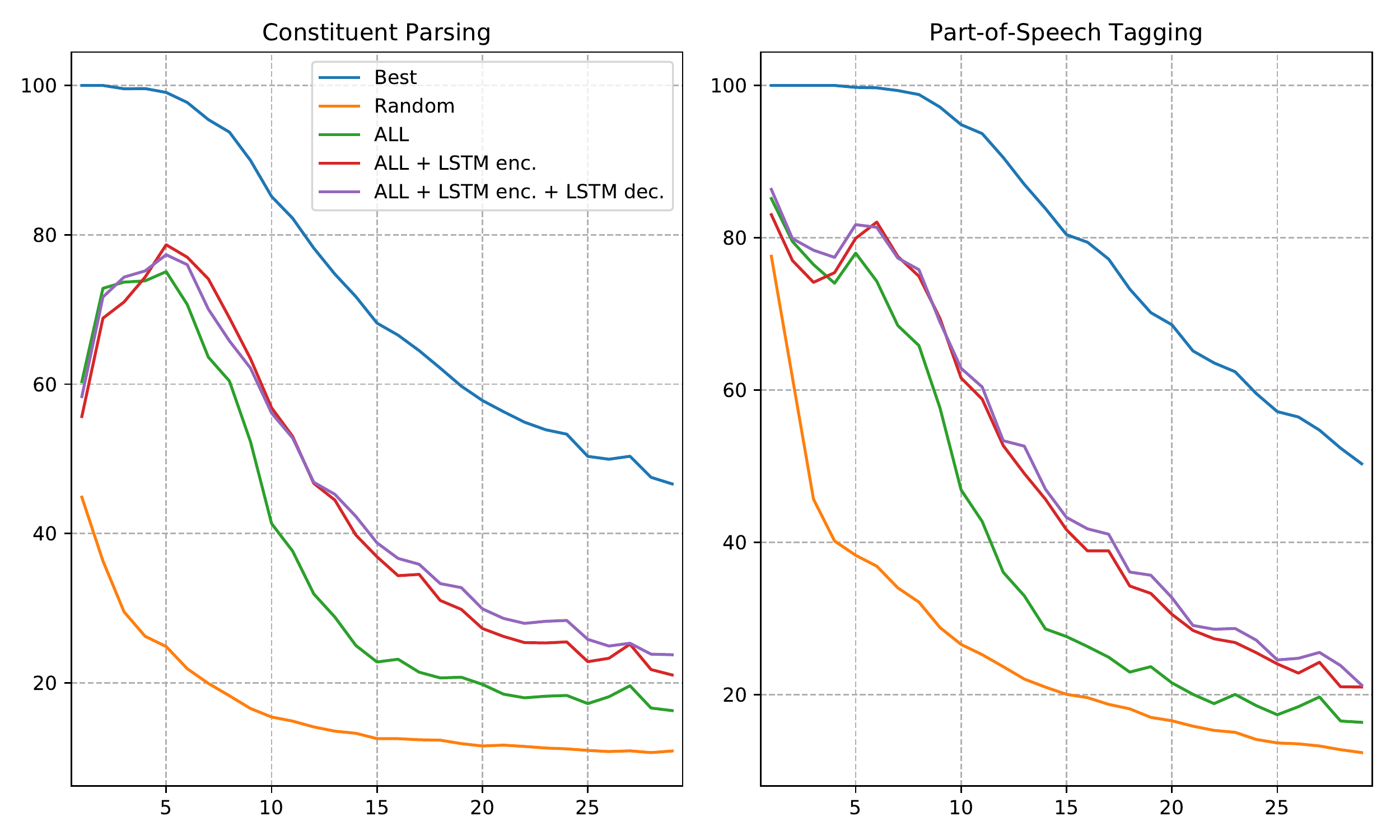}
    \caption{Constituency parsing $F_1$ scores and part-of-speech tagging accuracies by sentence length, for 1-nearest neighbor parsers based on semantic and syntactic variables, as well as a random baseline and an oracle nearest neighbor parser (``Best'').}
    \label{fig:length-breakdown}
\end{figure}

To investigate the performance gap between the bag-of-words \vg and \vg with LSTM modules, in Figure~\ref{fig:length-breakdown} we plot the performance of our models and baselines as the length of the target sentence increases. We see that performance in all settings degrades as the sentences get longer. This may be due to the fact that the data is much sparser as sentence length increases (leaving fewer candidate nearest neighbors for prediction). We also see that above 4 words or so the performance gap between the bag-of-words \vg and \vg with LSTM modules becomes more and more obvious. This may be because the bag-of-words encoder has a harder time capturing syntactic information as sentence length increases. In addition, there is a slight improvement from using an LSTM decoder when the sentence length increases beyond 12 or so, which suggests that a bag-of-words decoder may struggle to capture certain parts of the syntactic information in the sentence, even when using an LSTM encoder.

\begin{table*}[t]
\small
\setlength{\tabcolsep}{6pt}
\centering
\begin{tabular}{l|l}
\hline
\multirow{2}{*}{starting} & \textit{syntactic:} getting heading sitting chasing taking require trying sharing bothering pushing paying \\
& \textit{semantic:} begin start stopping forward rising wake initial starts goes started again getting beginning\\
\hline
\multirow{2}{*}{area} & \textit{syntactic:} engines certificate guests bottle responsibility lesson pieces suit bags vessel applications\\
 & \textit{semantic:} sector location zone fields rooms field places yard warehouse seats coordinates territory\\ 
\hline
\multirow{2}{*}{considered} & \textit{syntactic:} stable limited odd scary classified concerned awful purple impressive embarrassing jealous\\
& \textit{semantic:} thought assumed regard reasons wished understood purposes seemed expect guessed meant\\
\hline
\multirow{2}{*}{jokes} & \textit{syntactic:} gentlemen photos finding baby missile dna parent shop murder science recognition sheriff\\
& \textit{semantic:} funny humor prize stars cookie paradise dessert worthy smile happiness thrilled ideal kidding\\
\hline
\multirow{2}{*}{times} & \textit{syntactic:} princess officer wounds plan gang ships feelings user liar elements coincidence degrees pattern \\
& \textit{semantic:} twice later thousand pages seven every once often decade forgotten series four eight day time\\
\hline

\end{tabular}
\caption{Examples of the most similar words to particular query words using syntactic variable (first row) or semantic variable (second row). 
}
\label{lex-examples}
\end{table*}

\begin{table*}[t]
\small
\setlength{\tabcolsep}{2pt}
\centering
\begin{tabular}{|p{0.34\textwidth}|p{0.31\textwidth}|p{0.32\textwidth}|}
\hline
\multicolumn{1}{|c|}{Query Sentence} & \multicolumn{1}{c|}{Semantically Similar} & \multicolumn{1}{c|}{Syntactically Similar} \\
\hline
\hline
i have much more colours at home . & even if there was food , would n't it be at least 300 years old ? & you have a beautiful view from here . \\\hline
victor had never known darkness like it . & he had never experienced such darkness as this . & you seem like a really nice kid . \\\hline
this is , uh , too serious . & but this is too serious . & it is , however , illegal discrimination .\\\hline
you 're gon na save her life . & you will save her . & you 're gon na give a speech . \\\hline
we 've got to get a move on . & come on , we got ta move . & you 'll have to get in there . \\\hline
and that was usually the highlight of my day . & i really enjoyed it when i did it . & and yet that was not the strangest aspect of the painting . \\\hline
we do need to collect our taxes somehow . & we have to earn the money we need . & now i have to do my job .\\
\hline
this is just such a surprise . & oh . this is a surprise . & this is just a little gain . \\\hline
okay . aw , that 's so romantic . & it 's so romantic ! & oh . well , that 's not good . \\\hline
we 're gon na have to do something about this . & we 'll have to do something about that . & we 're gon na have to do something about yours . \\\hline
\end{tabular}
\caption{Examples of most similar sentences to particular query sentences in terms of the semantic variable or the syntactic variable. }
\label{sentence-examples}
\end{table*}

\subsection{Qualitative Analysis}
To qualitatively evaluate our latent variables, we find (via cosine similarity) nearest neighbor sentences to test set examples in terms of both the semantic and syntactic representations. We also find nearest neighbors of words (which we view as single-word sentences). We discuss the results of this analysis below.

\subsubsection{Lexical Analysis}

Table~\ref{lex-examples} shows word nearest neighbors for both syntactic and semantic representations. We see that the most similar words found by the syntactic variable share the same part-of-speech tags with the query words. For example, ``starting'' is close to ``getting'' and ``taking,''  even though these words are not semantically similar. Words retrieved according to the semantic variable, however, are more similar semantically, e.g., ``begin'' and ``starts''. As another example, ``times'' is similar to words that are either related to descriptions of frequency (e.g., ``twice'' and ``often'') or related to numbers (e.g., ``thousand'', ``seven'').

\subsubsection{Sentential Analysis}

As shown in Table~\ref{sentence-examples}, sentences that are similar in terms of their semantic variables tend to have similar semantics. However, sentences that are similar in terms of their syntactic variables are mostly semantically unrelated but have similar surface forms. For example, ``you 're gon na save her life .'' has the same meaning as ``you will save her .'' while having a similar syntactic structure to ``you 're gon na give a speech .'' (despite having very different meanings). As another example, although the semantic variable does not find a good match for ``i have much more colours at home .'', which can be attributed to the limited size of candidate sentences, the nearest syntactic neighbor (``you have a beautiful view from here .'') has a very similar syntactic structure to the query sentence.

\section{Discussion}

In this paper we explored simple methods to disentangle syntax and semantics in latent representations of sentences. One goal was to measure the impact of simple decisions on the disentanglement of both the semantic and syntactic variables, even when restricting ourselves to simplified bag-of-words encoders. Due to the constrained nature of these bag-of-words models, we found that it was important to use different word embedding spaces for the semantic and syntactic encoders. In preliminary experiments, we experimented with the use of the same word embedding space  but distinct feed-forward layers in the two latent variable encoders. However, this setting proved extremely difficult to achieve a disentanglement between syntax and semantics. Hence an important component of disentanglement with these bag-of-words encoders is the use of different word embedding spaces. 

We also conducted experiments using LSTM encoders and decoders as recurrent neural networks are a natural way to capture syntactic information in a sentence.
We found this approach to give us additional benefits for both disentangling semantics and syntax and achieving better results overall. Nonetheless, we find it encouraging that even when using bag-of-words encoders, our multi-task losses are able to achieve a separation as measured by our semantic and syntactic similarity tasks.

\section{Conclusion}
We proposed a generative model and several losses for disentangling syntax and semantics in sentence representations. We also proposed syntactic similarity tasks for measuring the amount of disentanglement between semantic and syntactic representations. We characterized the effects of the losses as well as the use of LSTM modules on both semantic tasks and syntactic tasks. Our models achieve the best performance across both sets of similarity tasks when the latent representations are most disentangled.

\section*{Acknowledgments}
We would like to thank the anonymous reviewers, NVIDIA for donating GPUs used in this research,
and Google for a faculty research award to K.~Gimpel that partially supported this research.

\bibliography{naaclhlt2019}

\begin{thebibliography}{61}
\expandafter\ifx\csname natexlab\endcsname\relax\def\natexlab#1{#1}\fi

\bibitem[{Agirre et~al.(2015)Agirre, Banea, Cardie, Cer, Diab, Gonzalez-Agirre,
  Guo, Lopez-Gazpio, Maritxalar, Mihalcea, Rigau, Uria, and
  Wiebe}]{agirre2015semeval}
Eneko Agirre, Carmen Banea, Claire Cardie, Daniel Cer, Mona Diab, Aitor
  Gonzalez-Agirre, Weiwei Guo, Inigo Lopez-Gazpio, Montse Maritxalar, Rada
  Mihalcea, German Rigau, Larraitz Uria, and Janyce Wiebe. 2015.
\newblock \href {https://doi.org/10.18653/v1/S15-2045} {{SemEval}-2015 task 2:
  Semantic textual similarity, {E}nglish, {S}panish and pilot on
  interpretability}.
\newblock In \emph{Proceedings of the 9th International Workshop on Semantic
  Evaluation (SemEval 2015)}, pages 252--263. Association for Computational
  Linguistics.

\bibitem[{Agirre et~al.(2014)Agirre, Banea, Cardie, Cer, Diab, Gonzalez-Agirre,
  Guo, Mihalcea, Rigau, and Wiebe}]{agirre2014semeval}
Eneko Agirre, Carmen Banea, Claire Cardie, Daniel Cer, Mona Diab, Aitor
  Gonzalez-Agirre, Weiwei Guo, Rada Mihalcea, German Rigau, and Janyce Wiebe.
  2014.
\newblock {SemEval}-2014 task 10: Multilingual semantic textual similarity.
\newblock In \emph{Proceedings of the 8th International Workshop on Semantic
  Evaluation (SemEval 2014)}, pages 81--91. Association for Computational
  Linguistics.

\bibitem[{Agirre et~al.(2016)Agirre, Banea, Cer, Diab, Gonzalez-Agirre,
  Mihalcea, Rigau, and Wiebe}]{agirre2016semeval}
Eneko Agirre, Carmen Banea, Daniel Cer, Mona Diab, Aitor Gonzalez-Agirre, Rada
  Mihalcea, German Rigau, and Janyce Wiebe. 2016.
\newblock \href {https://doi.org/10.18653/v1/S16-1081} {{SemEval}-2016 task 1:
  Semantic textual similarity, monolingual and cross-lingual evaluation}.
\newblock In \emph{Proceedings of the 10th International Workshop on Semantic
  Evaluation (SemEval-2016)}, pages 497--511. Association for Computational
  Linguistics.

\bibitem[{Agirre et~al.(2012)Agirre, Cer, Diab, and
  Gonzalez-Agirre}]{agirre2012semeval}
Eneko Agirre, Daniel Cer, Mona Diab, and Aitor Gonzalez-Agirre. 2012.
\newblock \href {http://aclweb.org/anthology/S12-1051} {{SemEval}-2012 task 6:
  A pilot on semantic textual similarity}.
\newblock In \emph{*SEM 2012: The First Joint Conference on Lexical and
  Computational Semantics -- Volume 1: Proceedings of the main conference and
  the shared task, and Volume 2: Proceedings of the Sixth International
  Workshop on Semantic Evaluation (SemEval 2012)}, pages 385--393. Association
  for Computational Linguistics.

\bibitem[{Agirre et~al.(2013)Agirre, Cer, Diab, Gonzalez-Agirre, and
  Guo}]{diab2013eneko}
Eneko Agirre, Daniel Cer, Mona Diab, Aitor Gonzalez-Agirre, and Weiwei Guo.
  2013.
\newblock \href {http://aclweb.org/anthology/S13-1004} {{*SEM} 2013 shared
  task: Semantic textual similarity}.
\newblock In \emph{Second Joint Conference on Lexical and Computational
  Semantics (*SEM), Volume 1: Proceedings of the Main Conference and the Shared
  Task: Semantic Textual Similarity}, pages 32--43. Association for
  Computational Linguistics.

\bibitem[{Alemi et~al.(2016)Alemi, Fischer, Dillon, and Murphy}]{alemi2016deep}
Alexander~A Alemi, Ian Fischer, Joshua~V Dillon, and Kevin Murphy. 2016.
\newblock Deep variational information bottleneck.
\newblock \emph{arXiv preprint arXiv:1612.00410}.

\bibitem[{Augenstein and S{\o}gaard(2017)}]{augen2017multitask}
Isabelle Augenstein and Anders S{\o}gaard. 2017.
\newblock \href {https://doi.org/10.18653/v1/P17-2054} {Multi-task learning of
  keyphrase boundary classification}.
\newblock In \emph{Proceedings of the 55th Annual Meeting of the Association
  for Computational Linguistics (Volume 2: Short Papers)}, pages 341--346.
  Association for Computational Linguistics.

\bibitem[{Bollmann et~al.(2018)Bollmann, S{\o}gaard, and
  Bingel}]{bollmann18multitask}
Marcel Bollmann, Anders S{\o}gaard, and Joachim Bingel. 2018.
\newblock \href {http://aclweb.org/anthology/W18-3403} {Multi-task learning for
  historical text normalization: Size matters}.
\newblock In \emph{Proceedings of the Workshop on Deep Learning Approaches for
  Low-Resource NLP}, pages 19--24. Association for Computational Linguistics.

\bibitem[{Bowman et~al.(2016)Bowman, Vilnis, Vinyals, Dai, Jozefowicz, and
  Bengio}]{bowman2016generating}
Samuel~R. Bowman, Luke Vilnis, Oriol Vinyals, Andrew Dai, Rafal Jozefowicz, and
  Samy Bengio. 2016.
\newblock \href {https://doi.org/10.18653/v1/K16-1002} {Generating sentences
  from a continuous space}.
\newblock In \emph{Proceedings of The 20th SIGNLL Conference on Computational
  Natural Language Learning}, pages 10--21. Association for Computational
  Linguistics.

\bibitem[{Cer et~al.(2017)Cer, Diab, Agirre, Lopez-Gazpio, and
  Specia}]{cer2017semeval}
Daniel Cer, Mona Diab, Eneko Agirre, Inigo Lopez-Gazpio, and Lucia Specia.
  2017.
\newblock \href {https://doi.org/10.18653/v1/S17-2001} {{SemEval}-2017 task 1:
  Semantic textual similarity multilingual and crosslingual focused
  evaluation}.
\newblock In \emph{Proceedings of the 11th International Workshop on Semantic
  Evaluation (SemEval-2017)}, pages 1--14. Association for Computational
  Linguistics.

\bibitem[{Chen et~al.(2018)Chen, Tang, Livescu, and Gimpel}]{chen2018vsl}
Mingda Chen, Qingming Tang, Karen Livescu, and Kevin Gimpel. 2018.
\newblock \href {http://aclweb.org/anthology/D18-1020} {Variational sequential
  labelers for semi-supervised learning}.
\newblock In \emph{Proceedings of the 2018 Conference on Empirical Methods in
  Natural Language Processing}, pages 215--226. Association for Computational
  Linguistics.

\bibitem[{Chen et~al.(2016)Chen, Duan, Houthooft, Schulman, Sutskever, and
  Abbeel}]{Chen16infogan}
Xi~Chen, Yan Duan, Rein Houthooft, John Schulman, Ilya Sutskever, and Pieter
  Abbeel. 2016.
\newblock \href {http://dl.acm.org/citation.cfm?id=3157096.3157340} {Infogan:
  Interpretable representation learning by information maximizing generative
  adversarial nets}.
\newblock In \emph{Proceedings of the 30th International Conference on Neural
  Information Processing Systems}, NIPS'16, pages 2180--2188, USA. Curran
  Associates Inc.

\bibitem[{Christodoulopoulos et~al.(2010)Christodoulopoulos, Goldwater, and
  Steedman}]{D10-1056}
Christos Christodoulopoulos, Sharon Goldwater, and Mark Steedman. 2010.
\newblock \href {http://aclweb.org/anthology/D10-1056} {Two decades of
  unsupervised pos induction: How far have we come?}
\newblock In \emph{Proceedings of the 2010 Conference on Empirical Methods in
  Natural Language Processing}, pages 575--584. Association for Computational
  Linguistics.

\bibitem[{Conneau et~al.(2017)Conneau, Kiela, Schwenk, Barrault, and
  Bordes}]{infersent}
Alexis Conneau, Douwe Kiela, Holger Schwenk, Lo{\"i}c Barrault, and Antoine
  Bordes. 2017.
\newblock \href {https://doi.org/10.18653/v1/D17-1070} {Supervised learning of
  universal sentence representations from natural language inference data}.
\newblock In \emph{Proceedings of the 2017 Conference on Empirical Methods in
  Natural Language Processing}, pages 670--680. Association for Computational
  Linguistics.

\bibitem[{Davidson et~al.(2018)Davidson, Falorsi, De~Cao, Kipf, and
  Tomczak}]{davidson2018hyperspherical}
Tim~R. Davidson, Luca Falorsi, Nicola De~Cao, Thomas Kipf, and Jakub~M.
  Tomczak. 2018.
\newblock Hyperspherical variational auto-encoders.
\newblock \emph{34th Conference on Uncertainty in Artificial Intelligence
  (UAI-18)}.

\bibitem[{Deudon(2018)}]{deudon2018learning}
Michel Deudon. 2018.
\newblock \href
  {http://papers.nips.cc/paper/7377-learning-semantic-similarity-in-a-continuous-space.pdf}
  {Learning semantic similarity in a continuous space}.
\newblock In S.~Bengio, H.~Wallach, H.~Larochelle, K.~Grauman, N.~Cesa-Bianchi,
  and R.~Garnett, editors, \emph{Advances in Neural Information Processing
  Systems 31}, pages 993--1004. Curran Associates, Inc.

\bibitem[{Devlin et~al.(2018)Devlin, Chang, Lee, and
  Toutanova}]{devlin2018bert}
Jacob Devlin, Ming-Wei Chang, Kenton Lee, and Kristina Toutanova. 2018.
\newblock Bert: Pre-training of deep bidirectional transformers for language
  understanding.
\newblock \emph{arXiv preprint arXiv:1810.04805}.

\bibitem[{Du et~al.(2018)Du, Li, He, Xu, Bing, and Wang}]{du2018variational}
Jiachen Du, Wenjie Li, Yulan He, Ruifeng Xu, Lidong Bing, and Xuan Wang. 2018.
\newblock Variational autoregressive decoder for neural response generation.
\newblock In \emph{Proceedings of the 2018 Conference on Empirical Methods in
  Natural Language Processing}, pages 3154--3163.

\bibitem[{Fu et~al.(2018)Fu, Tan, Peng, Zhao, and Yan}]{fu2018style}
Zhenxin Fu, Xiaoye Tan, Nanyun Peng, Dongyan Zhao, and Rui Yan. 2018.
\newblock Style transfer in text: Exploration and evaluation.
\newblock In \emph{{A}{A}{A}{I}}.

\bibitem[{Goodfellow et~al.(2014)Goodfellow, Pouget-Abadie, Mirza, Xu,
  Warde-Farley, Ozair, Courville, and Bengio}]{goodfellow2014generative}
Ian Goodfellow, Jean Pouget-Abadie, Mehdi Mirza, Bing Xu, David Warde-Farley,
  Sherjil Ozair, Aaron Courville, and Yoshua Bengio. 2014.
\newblock Generative {A}dversarial {N}ets.
\newblock In \emph{Proceedings of NIPS}.

\bibitem[{Goyal et~al.(2017)Goyal, Sordoni, C{\^o}t{\'e}, Ke, and
  Bengio}]{goyal2017z}
Anirudh Goyal Alias~Parth Goyal, Alessandro Sordoni, Marc-Alexandre
  C{\^o}t{\'e}, Nan~Rosemary Ke, and Yoshua Bengio. 2017.
\newblock Z-forcing: Training stochastic recurrent networks.
\newblock In \emph{Advances in neural information processing systems}, pages
  6713--6723.

\bibitem[{Gulrajani et~al.(2017)Gulrajani, Ahmed, Arjovsky, and
  Vincent~Dumoulin}]{Gulrajani2017wgan}
Ishaan Gulrajani, Faruk Ahmed, Martin Arjovsky, and Aaron~Courville
  Vincent~Dumoulin. 2017.
\newblock {I}mproved {T}raining of {W}asserstein {GAN}s.
\newblock In \emph{Proceedings of NIPS}.

\bibitem[{Guu et~al.(2018)Guu, Hashimoto, Oren, and Liang}]{kelvin2018gen}
Kelvin Guu, Tatsunori~B. Hashimoto, Yonatan Oren, and Percy Liang. 2018.
\newblock \href {http://aclweb.org/anthology/Q18-1031} {Generating sentences by
  editing prototypes}.
\newblock \emph{Transactions of the Association for Computational Linguistics},
  6:437--450.

\bibitem[{Higgins et~al.(2016)Higgins, Matthey, Pal, Burgess, Glorot,
  Botvinick, Mohamed, and Lerchner}]{higgins2016beta}
Irina Higgins, Loic Matthey, Arka Pal, Christopher Burgess, Xavier Glorot,
  Matthew Botvinick, Shakir Mohamed, and Alexander Lerchner. 2016.
\newblock beta-vae: Learning basic visual concepts with a constrained
  variational framework.
\newblock In \emph{ICLR}.

\bibitem[{Hochreiter and Schmidhuber(1997)}]{hochreiter1997long}
Sepp Hochreiter and J{\"u}rgen Schmidhuber. 1997.
\newblock Long short-term memory.
\newblock \emph{Neural computation}, 9(8):1735--1780.

\bibitem[{Hsu et~al.(2017)Hsu, Zhang, and Glass}]{hsu17unsup}
Wei-Ning Hsu, Yu~Zhang, and James Glass. 2017.
\newblock \href
  {http://papers.nips.cc/paper/6784-unsupervised-learning-of-disentangled-and-interpretable-representations-from-sequential-data.pdf}
  {Unsupervised learning of disentangled and interpretable representations from
  sequential data}.
\newblock In I.~Guyon, U.~V. Luxburg, S.~Bengio, H.~Wallach, R.~Fergus,
  S.~Vishwanathan, and R.~Garnett, editors, \emph{Advances in Neural
  Information Processing Systems 30}, pages 1878--1889. Curran Associates, Inc.

\bibitem[{Hu et~al.(2017)Hu, Yang, Liang, Salakhutdinov, and
  Xing}]{hu17control}
Zhiting Hu, Zichao Yang, Xiaodan Liang, Ruslan Salakhutdinov, and Eric~P. Xing.
  2017.
\newblock \href {http://proceedings.mlr.press/v70/hu17e.html} {Toward
  controlled generation of text}.
\newblock In \emph{Proceedings of the 34th International Conference on Machine
  Learning}, volume~70 of \emph{Proceedings of Machine Learning Research},
  pages 1587--1596, International Convention Centre, Sydney, Australia. PMLR.

\bibitem[{Iyyer et~al.(2018)Iyyer, Wieting, Gimpel, and
  Zettlemoyer}]{iyyer2018adversarial}
Mohit Iyyer, John Wieting, Kevin Gimpel, and Luke Zettlemoyer. 2018.
\newblock \href {https://doi.org/10.18653/v1/N18-1170} {Adversarial example
  generation with syntactically controlled paraphrase networks}.
\newblock In \emph{Proceedings of the 2018 Conference of the North American
  Chapter of the Association for Computational Linguistics: Human Language
  Technologies, Volume 1 (Long Papers)}, pages 1875--1885. Association for
  Computational Linguistics.

\bibitem[{John et~al.(2018)John, Mou, Bahuleyan, and
  Vechtomova}]{john2018disentangled}
Vineet John, Lili Mou, Hareesh Bahuleyan, and Olga Vechtomova. 2018.
\newblock Disentangled representation learning for text style transfer.
\newblock \emph{arXiv preprint arXiv:1808.04339}.

\bibitem[{Jurafsky(1988)}]{C88-1057}
Daniel Jurafsky. 1988.
\newblock \href {http://aclweb.org/anthology/C88-1057} {Issues in relating
  syntax and semantics}.
\newblock In \emph{Coling Budapest 1988 Volume 1: International Conference on
  Computational Linguistics}.

\bibitem[{Kingma and Welling(2014)}]{Kingma2014}
Diederik~P. Kingma and Max Welling. 2014.
\newblock {A}uto-{E}ncoding {V}ariational {B}ayes.
\newblock In \emph{Proceedings of ICLR}.

\bibitem[{Kiros et~al.(2015)Kiros, Zhu, Salakhutdinov, Zemel, Urtasun,
  Torralba, and Fidler}]{skipthoughts}
Ryan Kiros, Yukun Zhu, Ruslan~R Salakhutdinov, Richard Zemel, Raquel Urtasun,
  Antonio Torralba, and Sanja Fidler. 2015.
\newblock \href {http://papers.nips.cc/paper/5950-skip-thought-vectors.pdf}
  {Skip-thought vectors}.
\newblock In C.~Cortes, N.~D. Lawrence, D.~D. Lee, M.~Sugiyama, and R.~Garnett,
  editors, \emph{Advances in Neural Information Processing Systems 28}, pages
  3294--3302. Curran Associates, Inc.

\bibitem[{Klein and Manning(2004)}]{P04-1061}
Dan Klein and Christopher Manning. 2004.
\newblock \href {http://aclweb.org/anthology/P04-1061} {Corpus-based induction
  of syntactic structure: Models of dependency and constituency}.
\newblock In \emph{Proceedings of the 42nd Annual Meeting of the Association
  for Computational Linguistics (ACL-04)}.

\bibitem[{Makhzani et~al.(2015)Makhzani, Shlens, Jaitly, Goodfellow, and
  Frey}]{Makhzani2015}
Alireza Makhzani, Jonathon Shlens, Navdeep Jaitly, Ian Goodfellow, and Brendan
  Frey. 2015.
\newblock {A}dversarial {A}utoencoders.
\newblock \emph{arXiv:1511.05644}.

\bibitem[{Manning et~al.(2014)Manning, Surdeanu, Bauer, Finkel, Bethard, and
  McClosky}]{manning2014stanford}
Christopher Manning, Mihai Surdeanu, John Bauer, Jenny Finkel, Steven Bethard,
  and David McClosky. 2014.
\newblock The stanford corenlp natural language processing toolkit.
\newblock In \emph{Proceedings of 52nd annual meeting of the association for
  computational linguistics: system demonstrations}, pages 55--60.

\bibitem[{Marcus et~al.(1993)Marcus, Marcinkiewicz, and
  Santorini}]{Marcus:1993:BLA:972470.972475}
Mitchell~P. Marcus, Mary~Ann Marcinkiewicz, and Beatrice Santorini. 1993.
\newblock \href {http://dl.acm.org/citation.cfm?id=972470.972475} {Building a
  large annotated corpus of english: The penn treebank}.
\newblock \emph{Comput. Linguist.}, 19(2):313--330.

\bibitem[{Mathieu et~al.(2016)Mathieu, Zhao, Zhao, Ramesh, Sprechmann, and
  LeCun}]{mathieu2016disentangling}
Michael~F Mathieu, Junbo~Jake Zhao, Junbo Zhao, Aditya Ramesh, Pablo
  Sprechmann, and Yann LeCun. 2016.
\newblock Disentangling factors of variation in deep representation using
  adversarial training.
\newblock In \emph{Advances in Neural Information Processing Systems}, pages
  5040--5048.

\bibitem[{Miao et~al.(2016)Miao, Yu, and Blunsom}]{miao2016neural}
Yishu Miao, Lei Yu, and Phil Blunsom. 2016.
\newblock \href {http://proceedings.mlr.press/v48/miao16.html} {Neural
  variational inference for text processing}.
\newblock In \emph{Proceedings of The 33rd International Conference on Machine
  Learning}, volume~48 of \emph{Proceedings of Machine Learning Research},
  pages 1727--1736, New York, New York, USA. PMLR.

\bibitem[{Mitchell(2016)}]{W16-1615}
Jeff Mitchell. 2016.
\newblock \href {https://doi.org/10.18653/v1/W16-1615} {Decomposing bilexical
  dependencies into semantic and syntactic vectors}.
\newblock In \emph{Proceedings of the 1st Workshop on Representation Learning
  for NLP}, pages 127--136. Association for Computational Linguistics.

\bibitem[{Mitchell and Steedman(2015)}]{P15-1126}
Jeff Mitchell and Mark Steedman. 2015.
\newblock \href {https://doi.org/10.3115/v1/P15-1126} {Orthogonality of syntax
  and semantics within distributional spaces}.
\newblock In \emph{Proceedings of the 53rd Annual Meeting of the Association
  for Computational Linguistics and the 7th International Joint Conference on
  Natural Language Processing (Volume 1: Long Papers)}, pages 1301--1310.
  Association for Computational Linguistics.

\bibitem[{Pennington et~al.(2014)Pennington, Socher, and
  Manning}]{pennington2014glove}
Jeffrey Pennington, Richard Socher, and Christopher~D. Manning. 2014.
\newblock {GloVe}: Global vectors for word representation.
\newblock In \emph{Empirical Methods in Natural Language Processing (EMNLP)},
  pages 1532--1543.

\bibitem[{Peters et~al.(2018)Peters, Neumann, Iyyer, Gardner, Clark, Lee, and
  Zettlemoyer}]{N18-1202}
Matthew Peters, Mark Neumann, Mohit Iyyer, Matt Gardner, Christopher Clark,
  Kenton Lee, and Luke Zettlemoyer. 2018.
\newblock \href {https://doi.org/10.18653/v1/N18-1202} {Deep contextualized
  word representations}.
\newblock In \emph{Proceedings of the 2018 Conference of the North American
  Chapter of the Association for Computational Linguistics: Human Language
  Technologies, Volume 1 (Long Papers)}, pages 2227--2237. Association for
  Computational Linguistics.

\bibitem[{Plank et~al.(2016)Plank, S{\o}gaard, and
  Goldberg}]{plank2016multilingual}
Barbara Plank, Anders S{\o}gaard, and Yoav Goldberg. 2016.
\newblock \href {https://doi.org/10.18653/v1/P16-2067} {Multilingual
  part-of-speech tagging with bidirectional long short-term memory models and
  auxiliary loss}.
\newblock In \emph{Proceedings of the 54th Annual Meeting of the Association
  for Computational Linguistics (Volume 2: Short Papers)}, pages 412--418.
  Association for Computational Linguistics.

\bibitem[{Reed et~al.(2014)Reed, Sohn, Zhang, and Lee}]{reed2014learning}
Scott Reed, Kihyuk Sohn, Yuting Zhang, and Honglak Lee. 2014.
\newblock Learning to disentangle factors of variation with manifold
  interaction.
\newblock In \emph{International Conference on Machine Learning}, pages
  1431--1439.

\bibitem[{Rei(2017)}]{rei2017semi}
Marek Rei. 2017.
\newblock \href {https://doi.org/10.18653/v1/P17-1194} {Semi-supervised
  multitask learning for sequence labeling}.
\newblock In \emph{Proceedings of the 55th Annual Meeting of the Association
  for Computational Linguistics (Volume 1: Long Papers)}, pages 2121--2130.
  Association for Computational Linguistics.

\bibitem[{Rezende et~al.(2014)Rezende, Mohamed, and Wierstra}]{Rezende2014}
Danilo~J. Rezende, Shakir Mohamed, and Daan Wierstra. 2014.
\newblock {S}tochastic {B}ackpropagation and {A}pproximate {I}nference in
  {D}eep {G}enerative {M}odels.
\newblock In \emph{Proceedings of ICML}.

\bibitem[{Serban et~al.(2017)Serban, Ororbia, Pineau, and
  Courville}]{serban2017piecewise}
Iulian~Vlad Serban, Alexander~G. Ororbia, Joelle Pineau, and Aaron Courville.
  2017.
\newblock \href {https://doi.org/10.18653/v1/D17-1043} {Piecewise latent
  variables for neural variational text processing}.
\newblock In \emph{Proceedings of the 2017 Conference on Empirical Methods in
  Natural Language Processing}, pages 422--432. Association for Computational
  Linguistics.

\bibitem[{Shen et~al.(2017)Shen, Lei, Barzilay, and Jaakkola}]{shen2017style}
Tianxiao Shen, Tao Lei, Regina Barzilay, and Tommi Jaakkola. 2017.
\newblock Style transfer from non-parallel text by cross-alignment.
\newblock In \emph{Advances in Neural Information Processing Systems}, pages
  6830--6841.

\bibitem[{Tenenbaum and Freeman(2000)}]{tenenbaum2000separating}
Joshua~B Tenenbaum and William~T Freeman. 2000.
\newblock Separating style and content with bilinear models.
\newblock \emph{Neural computation}, 12(6):1247--1283.

\bibitem[{{van Valin, Jr.}(2005)}]{van2005exploring}
Robert~D. {van Valin, Jr.} 2005.
\newblock \emph{Exploring the Syntax-Semantics Interface}.
\newblock Cambridge University Press.

\bibitem[{Wieting et~al.(2016)Wieting, Bansal, Gimpel, and
  Livescu}]{wieting-16-full}
John Wieting, Mohit Bansal, Kevin Gimpel, and Karen Livescu. 2016.
\newblock Towards universal paraphrastic sentence embeddings.
\newblock In \emph{Proceedings of International Conference on Learning
  Representations}.

\bibitem[{Wieting and Gimpel(2017)}]{wieting-gimpel:2017:Long}
John Wieting and Kevin Gimpel. 2017.
\newblock \href {http://aclweb.org/anthology/P17-1190} {Revisiting recurrent
  networks for paraphrastic sentence embeddings}.
\newblock In \emph{Proceedings of the 55th Annual Meeting of the Association
  for Computational Linguistics (Volume 1: Long Papers)}, pages 2078--2088,
  Vancouver, Canada. Association for Computational Linguistics.

\bibitem[{Wieting and Gimpel(2018)}]{para-nmt-acl-18}
John Wieting and Kevin Gimpel. 2018.
\newblock \href {http://aclweb.org/anthology/P18-1042} {{ParaNMT-50M}: Pushing
  the limits of paraphrastic sentence embeddings with millions of machine
  translations}.
\newblock In \emph{Proceedings of the 56th Annual Meeting of the Association
  for Computational Linguistics (Volume 1: Long Papers)}, pages 451--462.
  Association for Computational Linguistics.

\bibitem[{Wiseman et~al.(2018)Wiseman, Shieber, and
  Rush}]{wiseman2018templates}
Sam Wiseman, Stuart Shieber, and Alexander Rush. 2018.
\newblock \href {http://aclweb.org/anthology/D18-1356} {Learning neural
  templates for text generation}.
\newblock In \emph{Proceedings of the 2018 Conference on Empirical Methods in
  Natural Language Processing}, pages 3174--3187. Association for Computational
  Linguistics.

\bibitem[{Wu and Goodman(2018)}]{wu2018multimodal}
Mike Wu and Noah Goodman. 2018.
\newblock Multimodal generative models for scalable weakly-supervised learning.
\newblock In \emph{Advances in Neural Information Processing Systems}.

\bibitem[{Xu and Durrett(2018)}]{xu2018spherical}
Jiacheng Xu and Greg Durrett. 2018.
\newblock \href {http://aclweb.org/anthology/D18-1480} {Spherical latent spaces
  for stable variational autoencoders}.
\newblock In \emph{Proceedings of the 2018 Conference on Empirical Methods in
  Natural Language Processing}, pages 4503--4513. Association for Computational
  Linguistics.

\bibitem[{Yin et~al.(2018)Yin, Zhou, He, and Neubig}]{yin18struct}
Pengcheng Yin, Chunting Zhou, Junxian He, and Graham Neubig. 2018.
\newblock \href {http://aclweb.org/anthology/P18-1070} {Structvae:
  Tree-structured latent variable models for semi-supervised semantic parsing}.
\newblock In \emph{Proceedings of the 56th Annual Meeting of the Association
  for Computational Linguistics (Volume 1: Long Papers)}, pages 754--765.
  Association for Computational Linguistics.

\bibitem[{Zhang and Shasha(1989)}]{zhang1989simple}
Kaizhong Zhang and Dennis Shasha. 1989.
\newblock Simple fast algorithms for the editing distance between trees and
  related problems.
\newblock \emph{SIAM journal on computing}, 18(6):1245--1262.

\bibitem[{Zhao et~al.(2018)Zhao, Kim, Zhang, Rush, and
  LeCun}]{zhao2018adversarially}
Junbo Zhao, Yoon Kim, Kelly Zhang, Alexander~M Rush, and Yann LeCun. 2018.
\newblock {A}dversarially {R}egularized {A}utoencoders.
\newblock In \emph{Proceedings of ICML}.

\bibitem[{Zhao et~al.(2017)Zhao, Zhao, and Eskenazi}]{zhao2017learning}
Tiancheng Zhao, Ran Zhao, and Maxine Eskenazi. 2017.
\newblock Learning discourse-level diversity for neural dialog models using
  conditional variational autoencoders.
\newblock In \emph{Proceedings of the 55th Annual Meeting of the Association
  for Computational Linguistics (Volume 1: Long Papers)}, volume~1, pages
  654--664.

\bibitem[{Zhou and Neubig(2017)}]{zhou2017multi}
Chunting Zhou and Graham Neubig. 2017.
\newblock \href {https://doi.org/10.18653/v1/P17-1029} {Multi-space variational
  encoder-decoders for semi-supervised labeled sequence transduction}.
\newblock In \emph{Proceedings of the 55th Annual Meeting of the Association
  for Computational Linguistics (Volume 1: Long Papers)}, pages 310--320.
  Association for Computational Linguistics.

\end{thebibliography}
\bibliographystyle{acl_natbib}
\end{document}